# Neutron Imaging Enhancement for Lithium Batteries Based on Swin-Transformer and Image Fusion

Hansheng Ding[1], Fanxu Meng[1], Tian Zhang[1*]

([1] Northeast Normal University, Changchun 130024, Jilin,China

Corresponding author. E-mail: zhangt100@nenu.edu.cn)

**Abstract:**Neutron imaging is highly sensitive to light elements such as lithium and is therefore valuable for non-destructive testing (NDT) and analyzing internal structural evolution in lithium batteries. However, limited neutron-source flux and detector performance often result in images with mixed noise, low contrast, and blurred edges, hindering accurate assessment of battery internal states. To address these issues, this paper proposes a lithium-battery neutron image enhancement method combining an improved Swin-Transformer with spatial-domain image fusion. First, an improved Swin-Transformer incorporating Multi-Dconv Head Transposed Attention (MDTA) and Global Response Normalization (GRN) extracts and reconstructs image features to suppress high-density white spots and Gaussian-Poisson mixed noise. Second, guided filtering and a gradient-adaptive mask are used for selective edge enhancement, while Contrast Limited Adaptive Histogram Equalization (CLAHE) improves local contrast. Finally, a spatial-frequency (SF)-based local block-weighted fusion mechanism adaptively combines the edge-enhanced and contrast-equalized images to preserve microscopic details while suppressing noise. Experiments show that, compared with BM3D, DnCNN, and GAN, the proposed method achieves strong performance in full-reference and no-reference evaluations. Its best PSNR and SSIM reach 33.0517 dB and 0.9637, respectively, while the NR-IQA and Natural Image Quality Evaluator (NIQE) scores are as low as 0.1794 and 4.0974. The method reduces artifacts and over-smoothing, improving the visual fidelity and structural identifiability of lithium-battery neutron images.

## 1. Introduction

With the rapid advancement of new energy technologies, lithium-ion batteries have been widely applied across various fields due to their high energy density and long cycle life [1]. During the manufacturing, operation, and aging processes of a battery, the evolution of its internal electrode structure, electrolyte distribution, and interfacial states plays a decisive role in battery performance, lifespan, and safety. Consequently, achieving non-destructive, *in-situ* observation of the internal states of lithium batteries holds significant scientific and engineering value for understanding failure mechanisms, optimizing manufacturing processes, and establishing state-of-health (SOH) evaluation models. Among various non-destructive testing techniques, traditional X-ray imaging is sensitive to high-atomic-number (high-$Z$) materials but lacks sufficient resolving power for light elements. In contrast, neutron imaging possesses strong penetrating capability, and its attenuation coefficient is highly sensitive to light elements such as lithium (Li). It can effectively distinguish low-atomic-number (low-$Z$) materials inside the battery—such as lithium-containing compounds, binders, and separators—from heavy-element structures like metallic casings and current collectors, making it a powerful tool for investigating the dynamic processes of lithium-ion distribution, migration, and deposition inside lithium batteries [2–8].

However, in practical applications, limited by neutron source flux, restricted exposure time, and the physical Modulation Transfer Function (MTF) constraints of detectors [9–12], acquired lithium battery neutron images typically suffer from severe degradation. These images are often accompanied by intense mixed noise, extremely low contrast, and blurred microscopic edges. In particular, during the imaging process, direct bombardment of the CCD detector by high-energy γ-rays and scattered neutrons causes random, unevenly distributed, and irregularly shaped bright noise spots—known as white spots—to appear in the image [13–14]. This not only distorts the quantitative analysis of internal lithium distribution but also masks microscopic defects such as micro-cracks and gas bubbles, thereby weakening the safety early-warning capability and reliability of non-destructive testing. As illustrated in Figure 1, real neutron images exhibit significant mixed degradation and white spot noise.

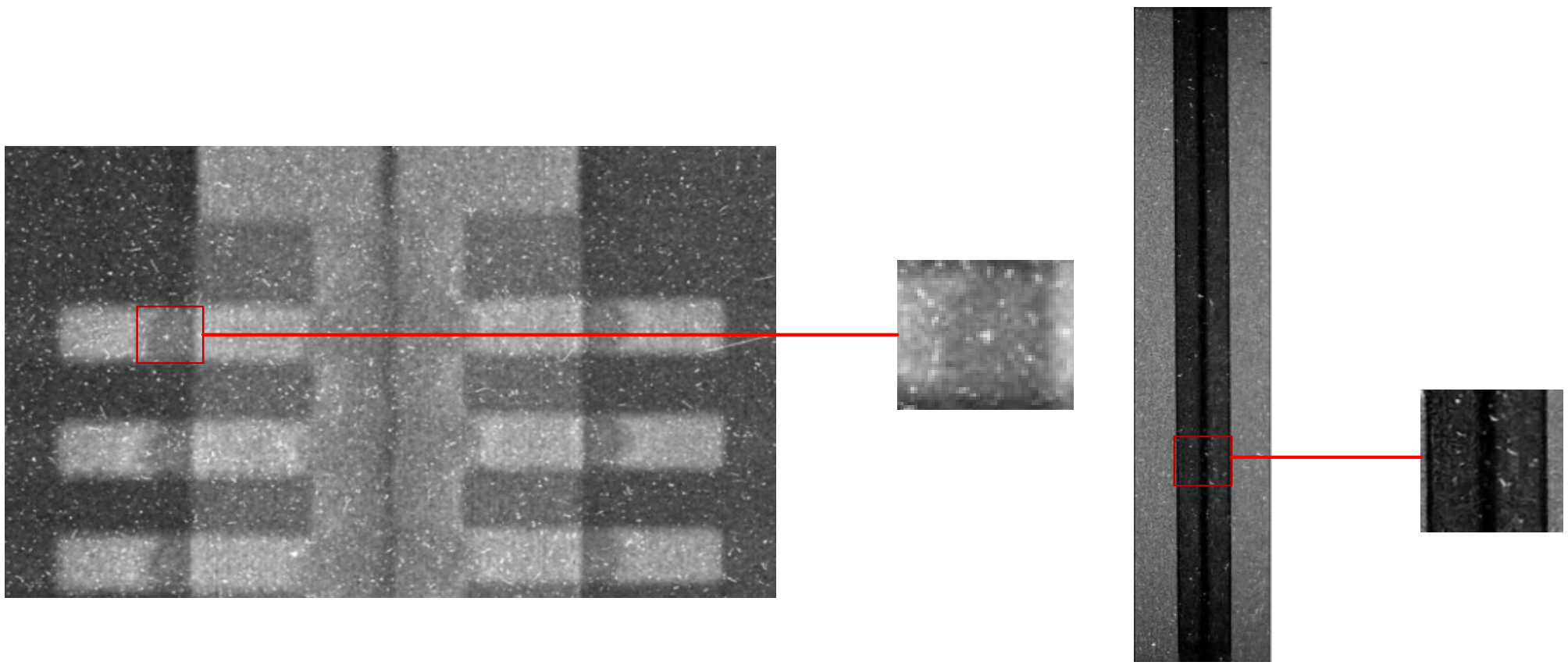

**Figure 1. Examples of mixed degradation and white spot noise in real neutron images.**

To address the issue where microscopic boundary details are easily lost during deep denoising of lithium battery neutron images, this paper proposes a cascaded enhancement framework combining improved Swin-Transformer restoration with spatial-domain multi-feature fusion. The main contributions of this study are summarized as follows:

**(1) Cascaded Hybrid Enhancement Architecture:** We propose a cascaded processing architecture of "deep denoising restoration—dual-branch feature enhancement—spatial-domain adaptive fusion". Through stage decoupling, it fully leverages the complementary advantages of deep networks in strongly suppressing complex mixed noise and traditional vision algorithms in reconstructing microscopic edge details with high fidelity.

**(2) Improved Feature Extraction Network:** By cascading the local window self-attention of Swin-Transformer with the Multi-Dconv Head Transposed Attention (MDTA) of Restormer [7], we achieve complementary integration of "local spatial–global channel" features. Furthermore, Global Response Normalization (GRN) [8] is introduced to enhance high-dimensional feature diversity, and a Dual Up-sample module is combined with global residual learning to effectively eliminate checkerboard artifacts and reduce optimization difficulty in image restoration.

**(3) High-Fidelity White Spot Noise Modeling and Dataset Construction:** We design a reverse extraction and synthesis method for real white spot noise based on Laplacian filtering, double-threshold segmentation, and binary flood-filling. Using this method, we construct a Lithium Battery Neutron Radiography Imaging (NRI) dataset containing over 10,000 "degraded–clean" sample pairs, making simulated degraded images closely match real physical imaging environments.

## 2. Related Work

Traditional methods have been widely introduced for neutron image restoration: Qiao et al. [18] applied the Total Variation (TV) model to neutron image denoising and adopted a hierarchical strategy to enhance details and avoid halo phenomena; Kazantsev et al. [19] proposed a reconstruction method combining Student's-*t* distribution data fitting with a TV penalty to eliminate stripe and ring artifacts; Zhao et al. [20] proposed an improved Robust Principal Component Analysis (RPCA) algorithm, establishing a "detection–localization–removal" mechanism to progressively eliminate white spot noise; Cao et al. [21] designed a weighted median filtering algorithm with dynamic detection and adaptive window expansion; specifically targeting lithium battery neutron projection images, Tan et al. [4] utilized median filtering combined with gradient operator sharpening and brightness adjustment to improve image quality.

However, traditional filtering algorithms typically rely on specific mathematical assumptions and struggle to balance denoising strength with structural preservation when facing high-density mixed random noise and scattering blur in real neutron imaging. In recent years, deep learning methods have become a research hotspot in neutron image restoration by virtue of their powerful nonlinear fitting capabilities and end-to-end multi-scale feature extraction mechanisms. Yang et al. [22] proposed a neutron image denoising and deblurring method based on Generative Adversarial Networks (GAN); Zharov et al. [23] employed a self-supervised learning strategy to learn image restoration from sample pairs with uncorrelated noise; Meng et al. [24] designed a multi-distortion suppression model based on GAN and constructed a neutron radiography dataset; Zhang et al. [25] combined SUNet with Monte

Carlo simulation to remove high-density white spots and Gaussian mixed noise; Li et al. [26] proposed a two-stage feature fusion denoising framework, DFGU-Net, specifically designed to suppress white spot noise.

Although deep learning has demonstrated potential in general neutron image denoising, dedicated restoration and enhancement algorithms for lithium battery neutron imaging remain scarce. Lithium batteries contain complex microscopic topological structures composed of tightly wound or stacked multi-layer positive and negative electrodes, current collectors, separators, and electrolytes [29–30]. Clear boundaries and interfaces (such as electrode/electrolyte wetting interfaces, electrode deformation contours, and micro-crack edges) serve as the core basis for evaluating lithium-ion transport kinetics, lithium plating phenomena, and local structural distortions [27–28, 32]. The inherent low-pass filtering characteristics of conventional deep learning denoising networks tend to smooth local pixels [3], easily causing "over-smoothing" artifacts along microstructural boundaries; conversely, directly applying traditional sharpening (such as Laplacian sharpening) to noisy images drastically amplifies residual noise, generating numerous spike artifacts on the gradient magnitude map (as shown in Figure 2). Therefore, there is an urgent need to design a collaborative enhancement algorithm capable of deeply filtering out mixed noise while selectively reinforcing internal microscopic boundaries and local contrast in batteries.

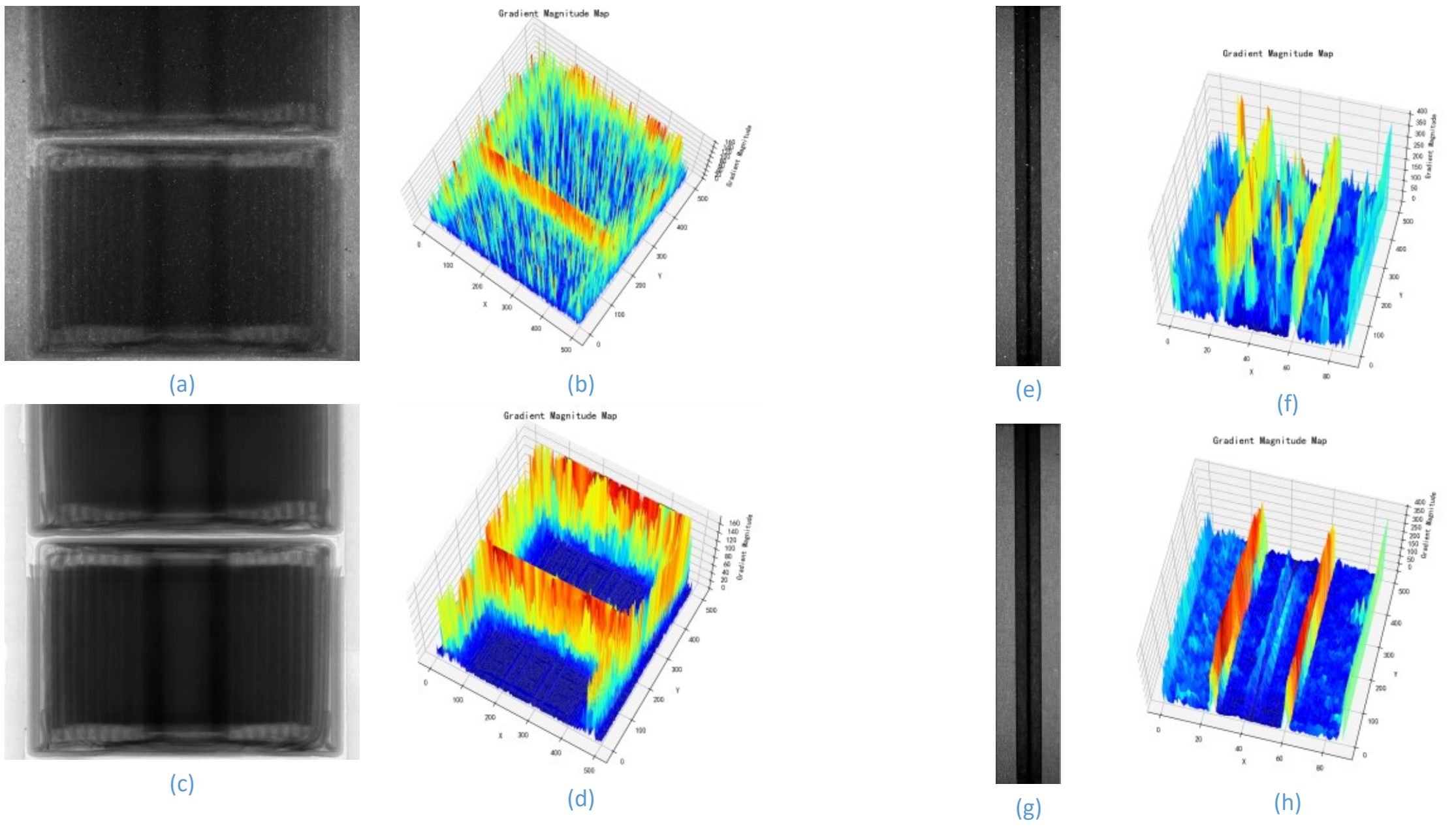

**Figure 2. Comparison of gradient magnitude surfaces before and after direct sharpening of a neutron image (direct sharpening in noisy regions generates numerous spike artifacts).**

## 3. Proposed Method

The proposed lithium battery neutron image enhancement algorithm based on Swin-Transformer and spatial-domain fusion comprises three core stages: First, the raw degraded image is fed into the improved Swin-Transformer restoration network to obtain a preliminarily denoised restored sub-image $I_1$. To compensate for the high-frequency boundary attenuation caused by deep denoising, $I_1$ subsequently undergoes two parallel enhancement branches: one branch employs selective edge enhancement based on guided filtering and a gradient-adaptive mask to extract and reinforce high-frequency structural information, yielding an edge-enhanced sub-image $I_2$; the other branch feeds $I_1$ into a Contrast Limited Adaptive Histogram Equalization (CLAHE) module to improve grayscale hierarchy, yielding a contrast-enhanced sub-image $I_3$. Finally, $I_2$ and $I_3$ are input into a spatial-frequency-based local block-weighted fusion module, which dynamically assigns pixel-block fusion weights according to local detail richness, ultimately outputting an enhanced image $I_{\text{enhanced}}$ characterized by suppressed noise, sharp edges, and balanced contrast.

## 3.1 Improved Swin-Transformer Restoration Network and Dataset Construction

### 3.1.1 Network Architecture

To fully exploit non-local self-similarity in neutron images and overcome the limited receptive field of traditional Convolutional Neural Networks (CNNs), we improve upon the Swin-Transformer-based U-Net architecture (SUNet) [6] and construct a restoration network integrating MDTA global channel attention [7], GRN global response normalization [8], and Dual Up-sample (the architecture is shown in Figure 3). The overall network adopts a symmetric Encoder–Decoder structure and a global residual learning strategy, primarily consisting of the following three modules:

**(1) Shallow Feature Extraction (SFE):** For an input degraded grayscale neutron image $Y$, a 3×3 convolutional layer $M_{\text{SFE}}(\cdot)$ is first utilized to extract low-frequency information and basic texture features $F_{\text{shallow}}$:

$$F_{shallow} = M_{SFE}(Y) \tag{1}$$

**(2) Improved Deep Feature Extraction:** The shallow features are subsequently fed into the U-Net deep feature space $M_{\text{UFE}}(\cdot)$ comprising an encoder, a bottleneck layer, and a decoder. Addressing the limitation of the original SUNet's local window attention in modeling global channel dependencies, we cascade the MDTA global channel attention module onto the standard Swin Transformer Block (STB) and embed GRN normalization into the feed-forward network to enhance high-dimensional feature diversity, extracting high-level multi-scale semantic features $F_{\text{deep}}$:

$$F_{deep} = M_{UFE}(F_{shallow}) \tag{2}$$

Within each STB core unit, Window-based Multi-Head Self-Attention (W-MSA) and Shifted Window-based Multi-Head Self-Attention (SW-MSA) are alternated in pairs to achieve cross-window information interaction while maintaining linear computational complexity. Given the input feature $f^{L-1}$ at layer $L$ - 1, the computation across two consecutive Swin Transformer Layers (STLs) is formulated as:

$$f^L = W - MSA\left(LN(f^{L-1})\right) + f^{L-1}, f^L = MLP\left(LN(f^L)\right) + f^L \quad (3)$$

$$f^{L+1} = SW - MSA\left(LN(f^L)\right) + f^L, f^{L+1} = MLP\left(LN(f^{L+1})\right) + f^{L+1} \quad (4)$$

where LN(·) denotes Layer Normalization, and MLP(·) denotes a Multi-Layer Perceptron incorporating a GELU activation function.

**(3) Dual-Branch Up-Sampling and Residual Reconstruction:** During the encoder down-sampling (Patch Merging) stage, the network concatenates adjacent 2×2 feature patches and adjusts the channel dimension via a linear layer to expand the receptive field. During the decoding stage, to avoid checkerboard artifacts caused by traditional transposed convolutions, a parallel Dual Up-sample module is adopted, simultaneously utilizing Bilinear interpolation and Pixel Shuffle branches for up-sampling; after passing through a 3×3 convolution and PReLU activation respectively, the two branches are concatenated and fused along the channel dimension. Finally, a 3×3 reconstruction convolutional layer $M_R(\cdot)$ maps the deep features back to the image space and adds them to the input image via a residual connection, outputting the preliminarily restored sub-image $I_1$: $I_1 = M_R(F_{deep}) + Y$.

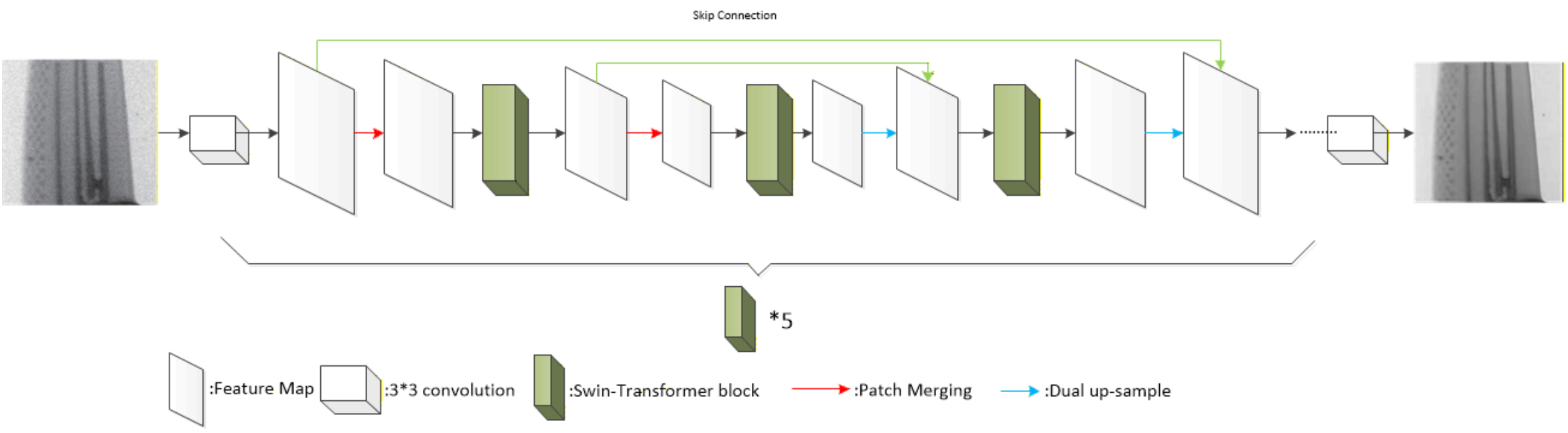

**Figure 3. Architecture flowchart of the improved Swin-Transformer restoration network.**

### 3.1.2 Degradation Modeling and White Spot Noise Dataset Construction

To address the scarcity of public high-definition lithium battery neutron image datasets, we select 40 high-resolution, low-noise real neutron images as clean references (Ground Truth) and introduce a multi-stage physical degradation model to synthesize training samples. First, Gaussian blur and disk defocus blur (parameterized by a Gaussian kernel and a circular point spread function $h_R$ of radius $R$, respectively) are sequentially applied to the clean reference image $I_{orig}$ to simulate the MTF limitations of the detector system and scattering diffusion effects; subsequently, zero-mean Gaussian noise $\eta \sim N(0, \sigma^2)$ with variance $\sigma^2$ and Poisson shot noise Poisson(·) are added to obtain the baseline degraded image $I_{deg}$.

On this basis, addressing the issues that traditional GAN generation [24] is prone to artifacts and Monte Carlo simulation [25] relies heavily on empirical parameters while struggling to reflect realistic non-uniform morphologies, we propose a real white spot noise extraction and synthesis method based on Laplacian filtering, double-threshold segmentation, and binary flood-filling. Let $I(x, y)$ denote a raw neutron image containing real white spots. First, an isotropic second-order differential Laplacian operator $K$ = [[0, -1, 0], [-1, 4, -1], [0, -1, 0]]$^T$ is applied to extract grayscale abrupt-change regions (using symmetric mirror reflection padding at boundaries to eliminate boundary effects). Since the edges of local high-brightness white spots exhibit strong positive and negative abrupt responses in the Laplacian response map $L(x, y)$, lower and upper thresholds $T_l$ = -225, $T_h$ = 225 are set for double-threshold segmentation to construct an initial binary edge mask $M_e(x, y)$ (where $M_e(x, y) = 1$ when $L(x, y) < T_l$ or $L(x, y) > T_h$, and 0 otherwise). To recover the complete interior regions of the white spots, a binary flood-fill operation Fill(·) is performed on the closed edge mask to obtain the interior region $M_f(x, y)$, which is then united with the edge region to isolate the pure white spot noise mask image $M_w(x, y)$.

Using this algorithm, a library of white spot samples is reversely extracted from a large collection of experimentally acquired neutron images. By performing random sampling and cropping (patch size 256×256) on full-size images and conducting pixel-wise truncated superposition of randomly cropped white spot noise maps $I_{\mathrm{spot}}(x, y)$ with baseline degraded images $I_{\mathrm{deg}}(x, y)$, we ultimately construct a high-fidelity training dataset dedicated to lithium battery neutron imaging containing over 10,000 "degraded–clean" image patch pairs.

### 3.2 Multi-Feature Image Enhancement and Spatial-Domain Weighted Fusion

#### 3.2.1 Selective Edge Enhancement Based on Guided Filtering and Gradient-Adaptive Mask

To prevent residual weak noise in flat regions of the denoised image $I_1$ from being amplified during edge sharpening, we adopt a selective edge enhancement strategy combining guided filtering with a gradient-adaptive mask. Using the restored image $I_1$ itself as the guidance image, a guided filtering operation GF(·) is performed to separate the smooth base layer $B$ = GF($I_1$, $I_1$) and the detail layer $D = I_1 - B$ containing high-frequency structures.

Subsequently, the Scharr operator, which offers superior directional sensitivity, is employed to compute the horizontal gradient $G_x$ and vertical gradient $G_y$ to obtain the gradient magnitude $G$. Based on the normalized gradient response $G_{\mathrm{norm}}(x, y)$, a Sigmoid adaptive edge weight mask $M(x, y) = 1 / [1 + exp(-k \cdot (G_{\mathrm{norm}}(x, y) - \tau))]$ is constructed, where $\tau$ is the gradient threshold controlling the edge selection range, and $k$ is the tuning parameter governing transition steepness. Applying this mask to the detail layer and superposing it back onto the restored image achieves selective reinforcement of structural edges, yielding the edge-enhanced sub-image $I_2(x, y) = I_1(x, y) + \alpha \cdot M(x, y) \cdot D(x, y)$ (where $\alpha$ is the detail enhancement gain factor).

#### 3.2.2 Contrast Limited Adaptive Histogram Equalization (CLAHE)

To address the narrow grayscale dynamic range in local regions and insufficient internal structural contrast caused by beam attenuation in neutron imaging, Contrast Limited Adaptive Histogram

Equalization (CLAHE) is introduced in parallel to process the restored sub-image $I_1$. CLAHE partitions the image into local tiles, clips the local histogram $H(r_k)$ using a preset threshold $T$ (i.e., $H_c(r_k) = \min(H(r_k), T)$) and uniformly redistributes the excess pixels across all grayscale levels, and subsequently completes local grayscale mapping via the Cumulative Distribution Function (CDF). This improves the grayscale hierarchy of the battery electrode and separator regions while suppressing excessive local noise amplification, outputting the contrast-enhanced sub-image $I_3$.

### 3.2.3 Spatial-Domain Weighted Fusion Based on Spatial Frequency

The edge-enhanced sub-image $I_2$ highlights high-frequency electrode boundaries and micro-crack contours inside the lithium battery, whereas the contrast-enhanced sub-image $I_3$ improves the overall grayscale distribution across local regions; the two exhibit strong informational complementarity. Therefore, we design a local block-adaptive weighted fusion strategy based on Spatial Frequency (SF).

First, sub-images $I_2$ and $I_3$ are partitioned into non-overlapping image blocks $P$ of size $M \times N$ (set to 8×8 in this study). For any image block, its horizontal row frequency $RF$, vertical column frequency $CF$, and overall spatial frequency $SF$ are computed respectively as:

$$RF = \sqrt{\frac{1}{M(N-1)}\sum_{i=1}^{M}\sum_{j=1}^{N-1}\big(P(i,j+1) - P(i,j)\big)^2} \tag{5}$$

$$CF = \sqrt{\frac{1}{(M-1)N}\sum_{i=1}^{M-1}\sum_{j=1}^{N}\big(P(i+1,j) - P(i,j)\big)^2} \tag{6}$$

$$SF = \sqrt{RF^2 + CF^2} + \epsilon \tag{7}$$

where $\varepsilon$ is a tiny constant to prevent division by zero. A larger $SF$ value indicates more intense gradient variations within the local block and richer effective edge and texture details. According to the spatial frequencies $SF_2$ and $SF_3$ of co-located blocks in the two sub-images, adaptive fusion weights $w_2$ and $w_3$ are dynamically assigned via normalization, and the local fused image block $P_{\text{fused}}$ is generated through weighted summation:

$$w_1 = \frac{SF_1}{SF_1 + SF_2} \tag{8}$$

$$w_2 = \frac{SF_2}{SF_1 + SF_2} \tag{9}$$

$$P_{fused} = w_1 \cdot P_1 + w_2 \cdot P_2 \tag{10}$$

Traversing and stitching all image blocks yields the final enhanced image $I_{\text{enhanced}}$.

## 4. Experimental Results and Analysis

The experiments in this study were conducted on a 64-bit Windows operating system equipped with an AMD Ryzen 9 9950X 16-Core Processor (4.30 GHz), 128 GB of system RAM, and an NVIDIA GeForce RTX 5090 GPU. The experimental data were sourced from the Swiss Spallation Neutron Source (SINQ) at the Paul Scherrer Institut (PSI) [33] and the open Battery Imaging Library of the Royal Society [34], with additional data support and testing validation provided by the China Institute of Atomic Energy (CIAE).

### 4.1 Comparison of Image Denoising and Restoration Performance

To comprehensively evaluate the performance of the proposed restoration network, the classical traditional denoising algorithm BM3D and mainstream deep learning models DnCNN and GAN were selected as baseline methods for qualitative and quantitative comparisons. Figure 4 illustrates the visual denoising effects of different algorithms on real lithium battery neutron images. As can be seen, BM3D only filters out a portion of low-density noise, leaving numerous bright white spots in the image; although DnCNN and GAN suppress most white spot noise, their removal of larger continuous white spots remains incomplete, and they produce noticeable "over-smoothing" artifacts in the internal electrode regions of the battery, blurring microscopic boundaries. In contrast, our proposed method thoroughly eliminates high-density white spots and background shot noise while effectively avoiding over-smoothing, clearly preserving the structural boundaries between electrodes and separators inside the lithium battery.

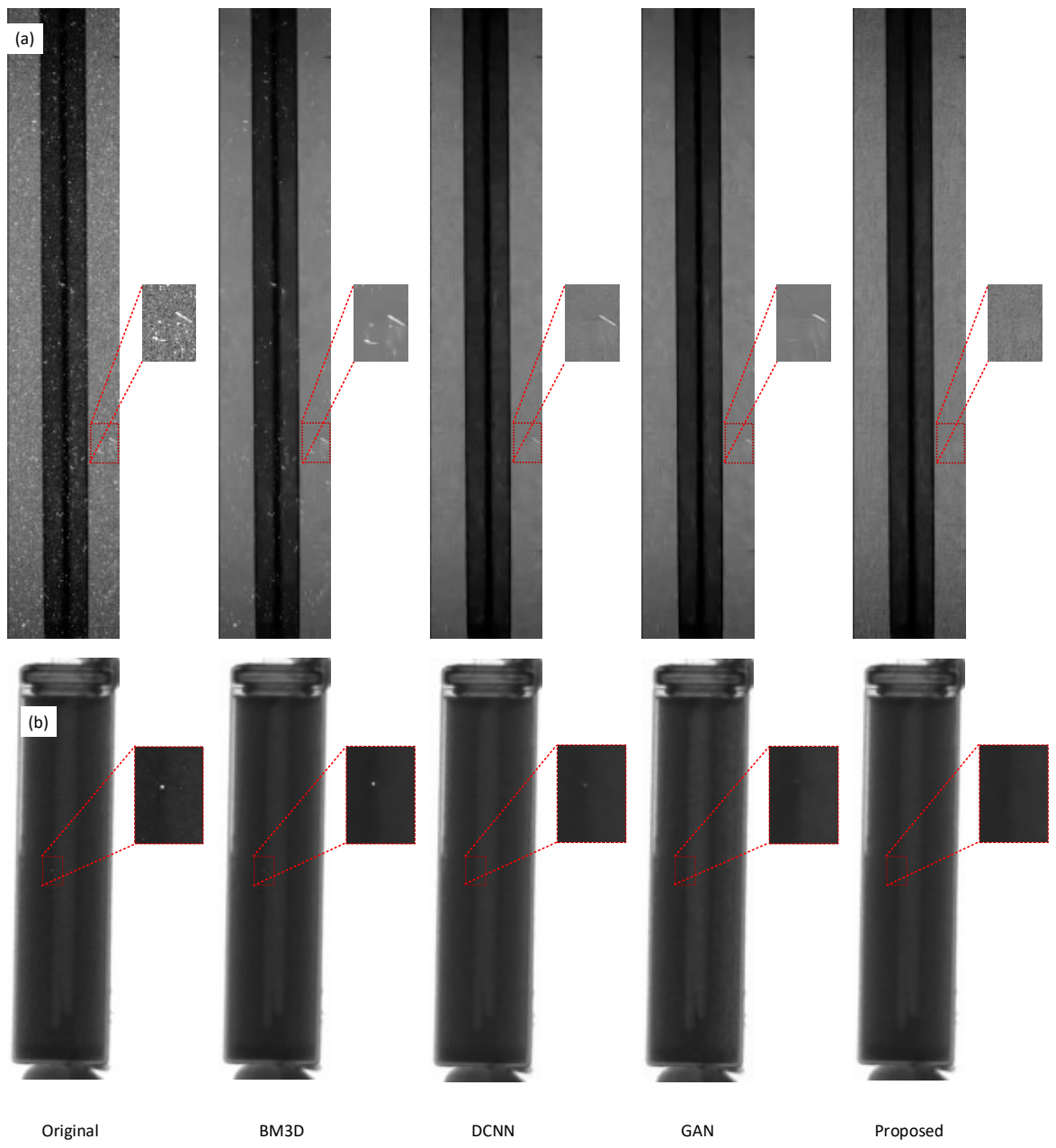

**Figure 4. Visual comparison of different denoising restoration algorithms on real lithium battery neutron**

**images: (a) Lithium battery image from the Swiss Spallation Neutron Source (SINQ) [33]; (b) Neutron NMC532 battery data [34].**

For quantitative evaluation, since experimentally acquired real neutron images (Figure 4(a) and (b)) lack noise-free ground truth references, we simultaneously introduce no-reference perceptual quality assessment metrics NR-IQA [35] and Natural Image Quality Evaluator (NIQE) [36] (for both metrics, lower values indicate superior perceptual quality), and utilize synthetically constructed paired degraded neutron images (Figure 5(c) and (d)) to conduct quantitative tests on full-reference metrics—Peak Signal-to-Noise Ratio (PSNR) and Structural Similarity (SSIM).

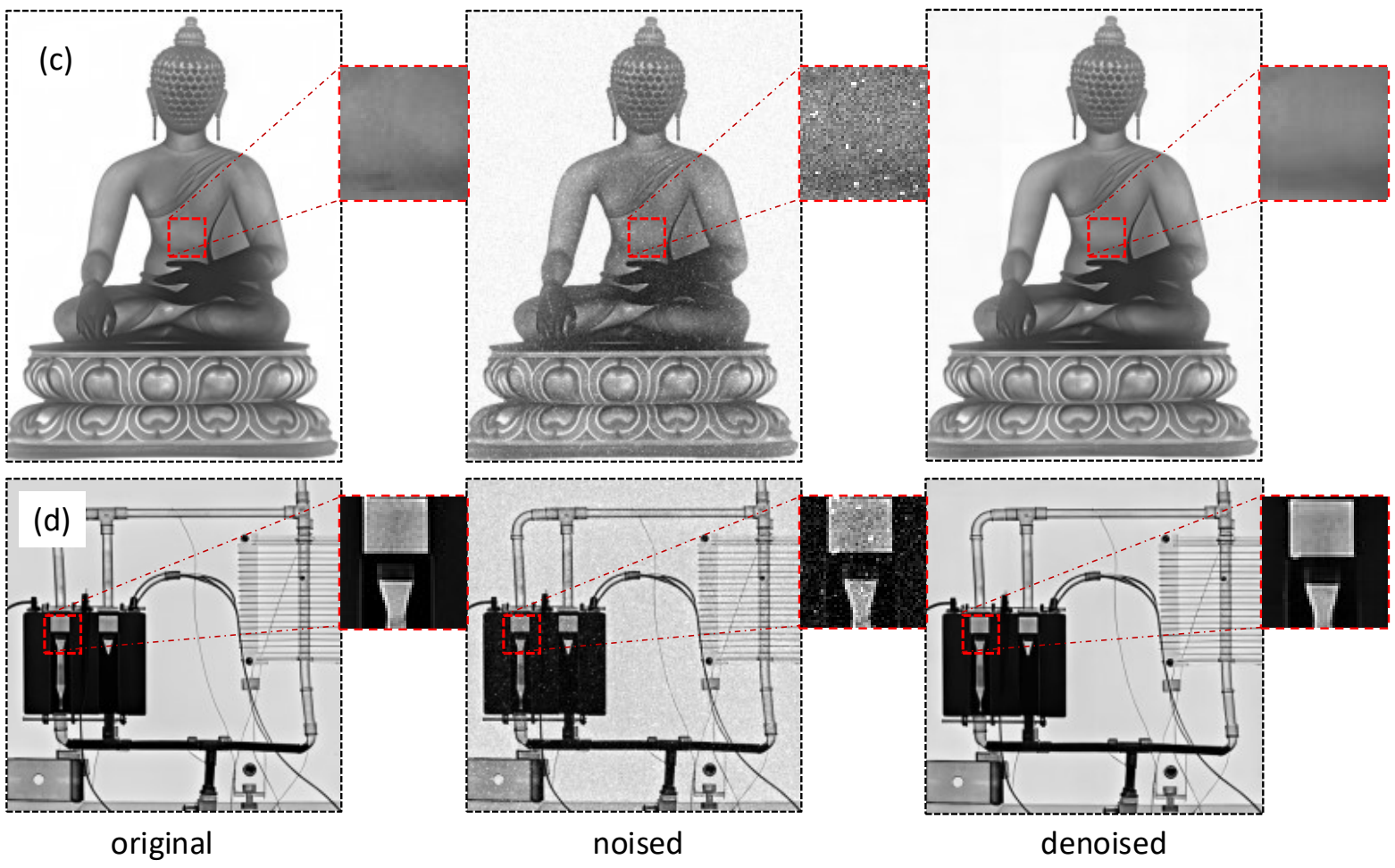

**Figure 5. Restoration comparison on synthetically constructed paired degraded neutron images (in (c) and (d), from left to right: original reference image, synthesized degraded image, and image restored by our proposed method).**

The objective evaluation metric comparisons of each denoising algorithm across the four test image sets are listed in Table 1.

**Table 1. Comparison of objective evaluation metrics across different denoising and restoration methods.**

| Metric | Test Image | Original | BM3D | DnCNN | GAN | Proposed |
|---|---|---|---|---|---|---|
| PSNR (dB) ↑ | (c) | — | 30.8058 | 32.5640 | 31.8431 | **32.7160** |
| PSNR (dB) ↑ | (d) | — | 30.8534 | 32.5369 | 31.9097 | **33.0517** |
| SSIM ↑ | (c) | — | 0.9392 | 0.9330 | 0.9345 | **0.9442** |
| SSIM ↑ | (d) | — | 0.9342 | 0.9479 | 0.9534 | **0.9637** |
| NR-IQA ↓ | (a) | 0.2021 | 0.1922 | 0.1863 | 0.1891 | **0.1794** |
| NR-IQA ↓ | (b) | 0.2381 | 0.2056 | 0.2153 | **0.1908** | 0.1985 |
| NR-IQA ↓ | (c) | 0.3524 | 0.2421 | 0.2531 | 0.2328 | **0.2316** |
| NR-IQA ↓ | (d) | 0.2851 | 0.1935 | 0.1839 | 0.1855 | **0.1798** |
| NIQE ↓ | (a) | 14.2559 | 10.4588 | 7.7545 | 6.3860 | **5.9540** |
| NIQE ↓ | (b) | 11.8619 | 10.9946 | 9.8812 | 9.3365 | **8.7751** |
| NIQE ↓ | (c) | 4.5049 | 6.3085 | 4.7204 | 4.7071 | **4.0974** |
| NIQE ↓ | (d) | 8.8451 | 9.8913 | **5.6930** | 6.8418 | 8.8452 |

As shown by the data in Table 1, on the full-reference test sets (c) and (d), the proposed method (Proposed) achieves the highest PSNR and SSIM scores. Specifically, on image (d), PSNR reaches 33.0517 dB, outperforming the traditional BM3D algorithm by 2.20 dB and the second-best DnCNN algorithm by 0.51 dB; the SSIM metric reaches up to 0.9637, verifying the superiority of cascading MDTA global channel attention with the dual-branch up-sampling mechanism in restoring structural fidelity. Among the no-reference evaluation metrics, our method obtains the lowest (optimal) NR-IQA scores on images (a), (c), and (d) (dropping as low as 0.1794), and achieves the lowest NIQE scores on real lithium battery images (a) and (b) as well as synthetic image (c) (where image (a) drops substantially from the original 14.2559 to 5.9540, and image (c) drops to 4.0974), demonstrating that the restored neutron images possess excellent visual naturalness and signal-to-noise ratio.

### 4.2 Comparison of Image Enhancement and Fusion Performance

To further validate the effectiveness of the "selective edge enhancement and spatial-frequency weighted fusion" module within the cascaded architecture, the proposed method is compared against Gamma correction (γ transformation), the Multi-Scale Retinex enhancement algorithm (Retinex) [31], and Bilateral Filtering enhancement (BF) on real lithium battery images (a) and (b). Gray Mean Gradient (GMG), Information Entropy (H), and the Sobel-operator-based Tenengrad gradient function are selected as objective evaluation metrics (higher values across all three metrics indicate richer texture hierarchy, more abundant information content, and higher edge sharpness, respectively). The experimental results are listed in Table 2.

**Table 2. Comparison of objective evaluation metrics across different image enhancement methods.**

| Metric | Test Image | γ Correction | **Retinex** | **BF** | **Proposed** |
|---|---|---|---|---|---|
| GMG ↑ | (a) | 8.0628 | 8.0948 | 2.0039 | **13.0827** |
| GMG ↑ | (b) | 1.5243 | 2.5034 | 1.1389 | **3.6697** |
| Entropy H ↑ | (a) | 6.6108 | 6.5951 | 6.3579 | **7.2184** |
| Entropy H ↑ | (b) | 6.1606 | 6.8781 | 6.2275 | **6.8965** |
| Tenengrad ↑ | (a) | 8128.4396 | **12578.7203** | 2372.1794 | 11174.7576 |
| Tenengrad ↑ | (b) | 441.2900 | 787.5873 | 758.7660 | **1399.9933** |

As evidenced by the data in Table 2, the proposed method significantly outperforms the baseline algorithms on the vast majority of metrics. On the Gray Mean Gradient (GMG) metric, our method reaches 13.0827 and 3.6697 on images (a) and (b), representing improvements of 61.6% and 46.6% over the second-best Retinex algorithm, respectively, proving that guided filtering combined with the gradient-adaptive mask greatly enriches the hierarchy of local microscopic edges. On the Information Entropy (H) metric, our method ranks first on both image sets (reaching 7.2184 and 6.8965, respectively), indicating that spatial-frequency weighted fusion thoroughly integrates the contrast information from the CLAHE branch and broadens the effective grayscale distribution range of the image. On the Tenengrad sharpness metric, our method achieves optimal performance on image (b) at 1399.9933 (a 77.8% improvement over Retinex); on image (a), it is slightly lower than Retinex because the Retinex algorithm indiscriminately amplifies residual local background noise when brightening dark regions without constraints, resulting in an artificially inflated Tenengrad gradient, whereas our method suppresses noise amplification in smooth regions via the Sigmoid gradient-adaptive mask, achieving superior visual fidelity while maintaining high sharpness.

## 5. Conclusion

Addressing the challenges of high-density gamma white spot noise, low contrast, and microstructural edge over-smoothing easily induced by deep denoising networks during lithium battery neutron imaging, this paper proposes a cascaded enhancement method based on an improved Swin-Transformer and spatial-domain image fusion. First, by combining second-order Laplacian differentiation, double-threshold segmentation, and binary flood-filling, white spot noise samples are accurately extracted from real neutron images to construct a high-fidelity lithium battery neutron image restoration dataset. Second, MDTA global channel attention, GRN normalization, and a dual-branch up-sampling mechanism are introduced into the SUNet architecture, achieving efficient filtering of complex mixed noise and high-fidelity structural reconstruction. Finally, dual branches of "guided-filtering gradient-adaptive edge enhancement" and "CLAHE contrast enhancement" are designed, and a local block-adaptive weighted fusion strategy based on spatial frequency is employed to achieve collaborative optimization of edge sharpness and local contrast. Quantitative and qualitative experiments on public datasets and real test data demonstrate that our proposed method significantly outperforms existing mainstream methods across PSNR (up to 33.0517 dB), SSIM (up to 0.9637), NR-IQA (down to 0.1794), NIQE (down to 4.0974), as well as post-enhancement GMG and Information Entropy metrics, providing a high-quality image enhancement tool for quantitative analysis of internal lithium-ion distribution and non-destructive microstructural testing in lithium batteries.